\documentclass[10pt,letterpaper]{article}
\usepackage[letterpaper,top=.73in,bottom=.73in,left=.82in,right=.82in]{geometry}
\usepackage[T1]{fontenc}
\usepackage[utf8]{inputenc}
\usepackage{newtxtext,newtxmath}
\usepackage{microtype}
\usepackage{graphicx,booktabs,tabularx,array}
\usepackage{amsmath}
\usepackage[hidelinks]{hyperref}
\usepackage{caption}
\usepackage{xcolor}
\hypersetup{pdfauthor={Jahnavi Yelamanchi},pdftitle={Reference Wording Shapes Retrieval of Robot Task Instructions}}
\begin{document}
\begin{center}
{\LARGE\bfseries Reference Wording Shapes Retrieval}\\[3pt]
{\LARGE\bfseries of Robot Task Instructions}\\[8pt]
{\normalsize Jahnavi Yelamanchi}
\end{center}
\vspace{-6pt}
\begin{abstract}
Identifying a robot task from a short instruction is sensitive to which descriptions are available for retrieval. We investigate this effect in the natural-language instructions included with the RLBench manipulation benchmark. We extract 296 fixed descriptions from 62 publicly available task files and compare five training-free lexical rankers over 80 random choices of one indexed description per task. Exact strings assigned to conflicting task labels are excluded from the primary score. Character $n$-gram TF--IDF achieves 61.3\% mean top-1 accuracy on the resulting held-out instructions, compared with 55.6\% for BM25 and 54.9\% for word TF--IDF. Across gallery choices, character TF--IDF ranges from 55.2\% to 66.8\%; its 5th--95th percentile range is 56.9\%--64.8\%. Adding a second indexed description improves character TF--IDF by 7.0 percentage points on matched queries. These results quantify the influence of reference wording and representation choice on instruction-to-task retrieval under a consistent evaluation protocol.
\end{abstract}
\vspace{-8pt}
\vspace{4pt}
\section{Introduction}
A language-conditioned system may need to select a task or skill from an instruction before any low-level control begins. The phrases ``open box'' and ``grasp the lid and open the box'' can describe the same action, whereas ``close box'' describes its reverse. A text retriever must use these differences without assuming a camera image or current robot state. Existing embodied systems go much further: CLIPort and PerAct use perceptual representations and control policies [5,6], while SayCan and RT-1 connect language to skill selection or action prediction [7,8]. We study that initial task-selection step, focusing on how the wording and number of reference descriptions affect retrieval accuracy.

RLBench provides task definitions with alternative natural-language descriptions [1,2]. These source-authored descriptions allow a controlled comparison of gallery composition, lexical representations, and wording ambiguity. Rather than indexing the first description in each task file, we resample the indexed descriptions and test retrieval on the remaining paraphrases.

This evaluation isolates instruction-to-task matching as one component of a language-conditioned robotics pipeline. Studying the retrieval step on its own makes differences due to reference wording visible before adding perception, state estimation, and robot control.
\begin{figure}[h!]
\centering\includegraphics[width=.96\linewidth]{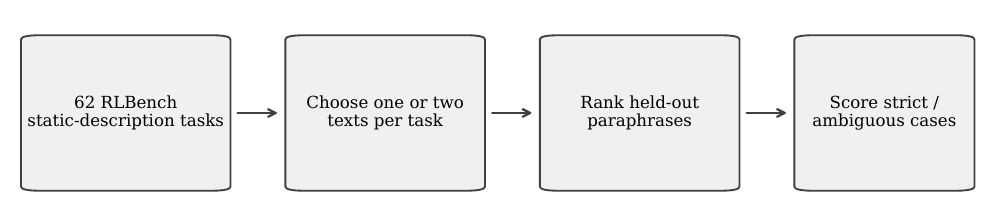}
\caption{Evaluation design. Existing RLBench instructions form task-wise reference galleries and held-out queries; the reference wording is resampled across trials.}
\end{figure}
\section{Data and evaluation}
We transcribed fixed, literal instructions from \texttt{init\_episode} task definitions in the public RLBench repository [2]. The corpus contains \textbf{62 task labels and 296 descriptions}; tasks have at least three usable descriptions each. Of 106 task definitions in the inspected repository directory, 62 yielded at least three fixed descriptions under this extraction protocol. The remaining 44 include runtime-generated templates and other cases requiring additional parsing. We preserve the source wording, including typographical errors.
\newpage
\section{Protocol and methods}
For each of 80 trials, a pseudorandom permutation is drawn separately for each task (NumPy generator seed 413). The first description in that trial is placed in the gallery; all remaining descriptions serve as queries. This produces 62 gallery entries and 234 candidate queries per trial. Each query is labeled by its source task. Retrieval scores depend on the selected gallery, so the reported average measures sensitivity to the available paraphrase, not an independent set of task episodes.

Two task definitions contain the identical description ``press the light switch'': \texttt{lamp\_on} and \texttt{lamp\_off}. For text-only retrieval, the gold task cannot be uniquely determined from that phrase. We therefore exclude instances of this \emph{exact cross-task duplicate} from primary accuracy (between zero and two queries are excluded per draw). Other ambiguous-looking but non-identical strings stay in the evaluation. The code additionally reports strict one-label and permissible-label scores for auditing.

For query $q$ and task $t$, with $G_t$ the set of indexed descriptions of task $t$, we score
\begin{equation}
S(t,q) = \max_{d \in G_t}s(q,d),\qquad
\widehat{t}(q)=\underset{t\in\mathcal{T}}{\arg\max}\,S(t,q).
\end{equation}
One-description galleries have $|G_t|=1$. Rankings use deterministic tie ordering, and inverse-document-frequency and decomposition statistics are fitted only on the gallery text within each draw.

\begin{table}[h!]
\caption{Data accounting and comparison designs. Exclusion applies only to exact descriptions occurring under multiple task labels.}
\centering\small
\begin{tabularx}{.98\linewidth}{@{}Xrrr@{}}
\toprule
\textbf{Design} & \textbf{Tasks} & \textbf{Gallery} & \textbf{Query candidates}\\
\midrule
One-description, 80 trials & 62 & 62 & 234 per trial\\
Two-description matched comparison, 30 trials & 62 & 124 & descriptions at position $\geq3$\\
First-listed reference, audit only & 62 & 62 & 234\\
\bottomrule
\end{tabularx}
\end{table}

\subsection{Retrieval baselines}
\textbf{BM25} applies lowercase alphanumeric tokenization, $k_1=1.2$ and $b=0.75$, and positive smoothed inverse-document frequency, without tuning to this dataset [9]. \textbf{Word TF--IDF} uses unigram/bigram features, sublinear term frequency, and cosine similarity [10,12]. \textbf{Character TF--IDF} uses within-word character $n$-grams ($n=3$--$5$), which can remain shared under small word-form changes. \textbf{Word+Char} averages those two cosine similarity scores with equal weights. \textbf{LSA-32} reduces the gallery-fitted word TF--IDF matrix to at most 32 singular-vector components and ranks by cosine similarity [11]. All settings were fixed for this evaluation; there was no per-query tuning.

\subsection{Gallery sensitivity and paired query design}
Across 80 trials, we summarize mean and sample standard deviation of top-1 accuracy, observed minimum and maximum, and the 5th and 95th percentiles. These statistics summarize \emph{variation across reference galleries on the same corpus}; they are distinct from population confidence intervals. Method comparisons are paired by trial. A second analysis repeats 30 of those gallery selections with two descriptions per task. In both conditions, only query descriptions at shuffled position three or later are evaluated, so adding the second example cannot make a previously evaluated query disappear from only one condition.

\begin{figure}[h!]
\centering\includegraphics[width=.70\linewidth]{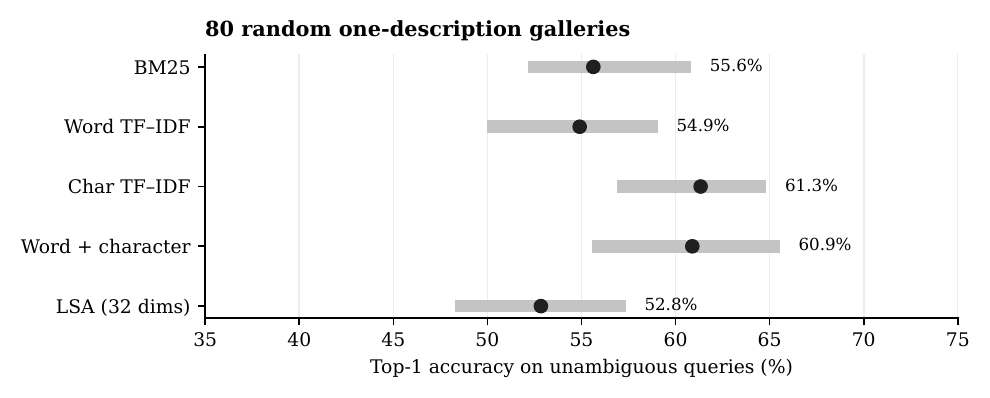}
\caption{Mean top-1 accuracy (dots) and 5th--95th percentile range (thick lines) over 80 random galleries. The whiskers show gallery-choice variability, not confidence bounds.}
\end{figure}
\newpage
\section{Retrieval results}
\begin{table}[h!]
\caption{One-description results over 80 gallery choices. Scores are percentages on unambiguous query texts. SD is across gallery draws; min--max is the observed range.}
\centering\small
\begin{tabularx}{\linewidth}{@{}Xrrrr@{}}
\toprule
\textbf{Ranker} & \textbf{Mean} & \textbf{SD} & \textbf{5th--95th} & \textbf{Min--max}\\
\midrule
BM25 & 55.6 & 2.5 & 52.1--60.8 & 51.1--62.1\\
Word TF--IDF & 54.9 & 3.0 & 50.0--59.1 & 47.8--61.6\\
Char TF--IDF & \textbf{61.3} & 2.6 & 56.9--64.8 & 55.2--66.8\\
Word+Char & 60.9 & 3.0 & 55.6--65.5 & 53.8--67.7\\
LSA-32 & 52.8 & 2.9 & 48.3--57.3 & 47.0--60.8\\
\bottomrule
\end{tabularx}
\end{table}

Character TF--IDF has the highest mean in Table 2, although Word+Char has the highest score in some individual galleries. The average paired advantage of character TF--IDF is 5.7 percentage points over BM25 and 6.4 over word TF--IDF. It outperforms BM25 in 79 of the 80 selected galleries (the remaining draw ties). The consistency across reference selections indicates that character-level features capture useful lexical variation in these task paraphrases.

The first-listed-gallery result would have obscured the sensitivity: on that one deterministic gallery, Word+Char reaches 63.8\% accuracy and character TF--IDF 62.9\%, but in repeated sampling their averages reverse. Benchmark reports that rely on a single arbitrary paraphrase can therefore change their ranking without changing the model code or task set.

\begin{figure}[h!]
\centering\includegraphics[width=.88\linewidth]{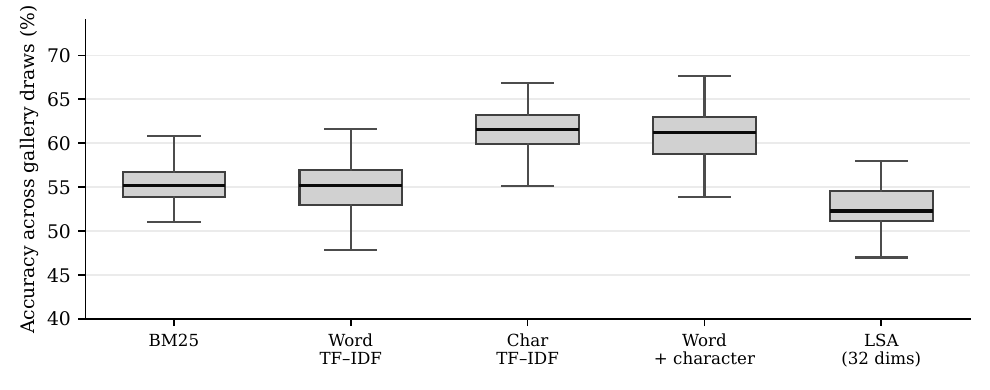}
\caption{Distribution of trial accuracies for the same five rankers. Boxes summarize variation attributable to which task descriptions were selected as references.}
\end{figure}

\subsection{Interpretation}
Short instructions create a strong surface-form effect. BM25 rewards shared informative terms; character $n$-grams can bridge inflections, spacing, and some spelling differences, but can also overemphasize shared object names in opposing commands. LSA discards some lexical detail, which need not be helpful when a single directional verb distinguishes two tasks. These patterns motivate targeted perturbation tests to distinguish the contributions of inflection, spelling, object names, and directional verbs.

The evaluation includes a known source artefact: some RLBench descriptions contain typographical errors or concatenated literals. We preserve them as authored, rather than manually correcting individual errors after examining model predictions. Preserving the original text also allows later work to measure how normalization or error correction changes retrieval.
\section{Indexing a second description}
To isolate gallery size from test-set composition, we evaluate descriptions from the third position onward under two index conditions: one or two earlier descriptions per task. Thirty permutations are sampled using the same random generator as the primary test. The held-out queries for each paired condition are identical within a trial, and both conditions exclude exact cross-task duplicate instructions.
\begin{figure}[h!]
\centering\includegraphics[width=.84\linewidth]{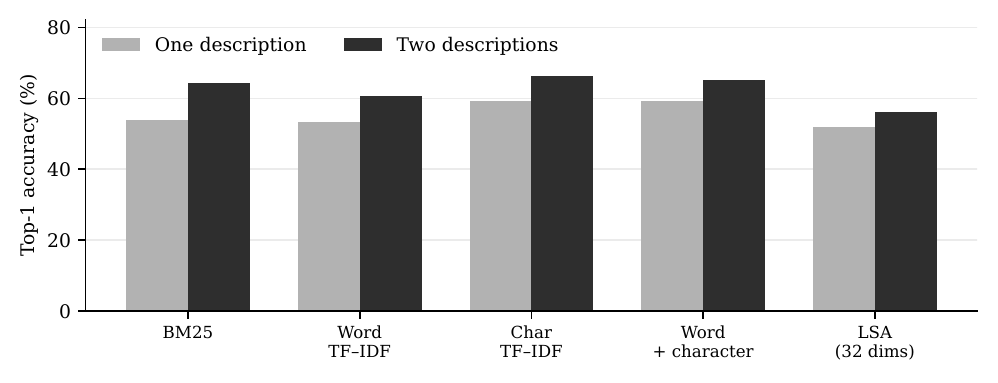}
\caption{Matched one- versus two-description galleries (30 trials). The candidate queries are unchanged between conditions within each trial.}
\end{figure}
\begin{table}[h!]
\caption{Average top-1 accuracy in the matched setting. The change is expressed in percentage points (pp).}
\centering\small
\begin{tabularx}{\linewidth}{@{}Xrrr@{}}
\toprule
\textbf{Ranker}&\textbf{One example}&\textbf{Two examples}&\textbf{Difference (pp)}\\
\midrule
BM25 &54.0\%&64.2\%&+10.2\\
Word TF--IDF&53.3\%&60.7\%&+7.4\\
Char TF--IDF&59.3\%&66.4\%&+7.0\\
Word+Char&59.2\%&65.1\%&+5.9\\
LSA-32&52.0\%&56.2\%&+4.3\\
\bottomrule
\end{tabularx}
\end{table}

The added description helps every ranker on average. BM25 has the largest increase, while character TF--IDF maintains the highest two-description mean. Because the one- and two-description scores are computed from the same queries, the result reflects additional indexed language, not the exclusion of a hard subset. This comparison isolates the value of a second available reference description under identical query conditions.

\section{Ambiguous task descriptions}
The description ``press the light switch'' occurs under both \texttt{lamp\_on} and \texttt{lamp\_off}. Because the target state is unspecified, the phrase alone does not identify a unique task label. We exclude these exact cross-task duplicates from primary accuracy and provide strict and either-label scores for comparison.

\newpage
\section{Relation to prior work}
RLBench supports learning and evaluation across manipulation tasks [1], while ALFRED [3] and CALVIN [4] connect natural-language instructions to grounded sequences or policies. CLIPort [5], PerAct [6], SayCan [7], and RT-1 [8] further combine language with perception, skill selection, or action prediction. Our experiment examines a complementary upstream question: how reliably task labels can be retrieved from alternative written instructions when the reference gallery itself varies.

The study compares established sparse and low-rank retrieval methods [9--12] under a shared, randomized reference-gallery protocol. Sentence-BERT [13] offers a natural follow-up for assessing semantic matching beyond token overlap. A script for \texttt{all-MiniLM-L6-v2} is included in the reproducibility package; neural sentence-embedding scores are not part of the reported results.

\subsection{Audit checks on scoring}
Exact cross-task duplicates are rare in this selected corpus, but their treatment should be visible. Over 80 draws, mean strict top-1 accuracy is 60.9\% for character TF--IDF and 55.4\% for BM25. Scoring either source label as correct for a duplicated phrase gives 61.0\% and 55.6\%, respectively; excluding duplicated query texts gives 61.3\% and 55.6\%. Thus the principal gap is not an artefact of dropping the switch instruction.

\begin{table}[h!]
\caption{Scoring audit: means over the same 80 one-description gallery draws. ``Either label'' accepts any original label for a repeated source string.}
\centering\small
\begin{tabularx}{\linewidth}{@{}Xrrr@{}}
\toprule
\textbf{Ranker} & \textbf{Strict} & \textbf{Either label} & \textbf{Duplicates removed}\\
\midrule
BM25 &55.4\%&55.6\%&55.6\%\\
Char TF--IDF &60.9\%&61.0\%&61.3\%\\
Word TF--IDF &54.7\%&55.1\%&54.9\%\\
\bottomrule
\end{tabularx}
\end{table}

A first-description-only gallery provides concrete examples of residual errors: BM25 retrieves \texttt{open\_door} for ``open the microwave door,'' while character TF--IDF retrieves \texttt{unplug\_charger} for ``plug the charger into the mains.'' These are verified outputs from the audit gallery, \emph{not} frequencies over all random galleries. Both mistakes collapse related actions despite distinctive wording.

\section{Discussion and conclusion}
The findings characterize written instruction matching on 296 fixed descriptions from 62 RLBench task files. The subset omits tasks with dynamic instruction generation, and queries consist of benchmark-authored paraphrases rather than independently collected operator requests. The repeated-gallery design captures sensitivity to indexed wording; extending it to all task definitions, pretrained embeddings, and embodied task execution would test how these patterns transfer to broader settings.

Two findings stand out. First, reference wording can alter the ordering of otherwise unchanged retrieval methods, making repeated gallery selections valuable for comparisons of short robot instructions. Second, adding one more reference description improves retrieval even when the evaluated queries stay fixed. The duplicate light-switch instruction further shows why label ambiguity deserves an explicit scoring policy. Together, these observations provide an empirical starting point for studying the language interface to larger robot-task libraries.

\section*{Reproducibility statement}
Results were computed with the attached Python code, NumPy and scikit-learn [14], with fixed pseudo-random seed 413. The package provides corpus task identifiers, the extracted description file for private verification, per-trial scores, generation scripts for all vector figures, and a separate sentence-transformer evaluation script. Source descriptions should be rechecked against their upstream files before public redistribution; the RLBench repository's custom licence [2] applies.

\vspace{3pt}
\noindent\textbf{Data availability.} The source of the task-description text is the public RLBench repository [2]. The full study package should only be publicly redistributed following review of the upstream licence terms.

\newpage
\section*{References}
\vspace{-3pt}
\begingroup\small\setlength{\parskip}{6.7pt}
\noindent [1] S. James, Z. Ma, D. Rovick Arrojo, and A. J. Davison. ``RLBench: The Robot Learning Benchmark \& Learning Environment.'' \emph{IEEE Robotics and Automation Letters}, 5(2), 2020, pp. 3019--3026. \href{https://doi.org/10.1109/LRA.2020.2974707}{DOI:10.1109/LRA.2020.2974707}.\par
\noindent [2] S. James and contributors. \emph{RLBench task source files}, \texttt{rlbench/tasks/}. Imperial College London; source revision \texttt{02720bba4c73}. \href{https://github.com/stepjam/RLBench}{Repository and licence}.\par
\noindent [3] M. Shridhar et al. ``ALFRED: A Benchmark for Interpreting Grounded Instructions for Everyday Tasks.'' \emph{CVPR}, 2020, pp. 10740--10749. \href{https://openaccess.thecvf.com/content_CVPR_2020/html/Shridhar_ALFRED_A_Benchmark_for_Interpreting_Grounded_Instructions_for_Everyday_Tasks_CVPR_2020_paper.html}{CVF}.\par
\noindent [4] O. Mees, L. Hermann, E. Rosete-Beas, and W. Burgard. ``CALVIN: A Benchmark for Language-Conditioned Policy Learning for Long-Horizon Robot Manipulation Tasks.'' \emph{IEEE Robotics and Automation Letters}, 7(3), 2022, pp. 7327--7334. \href{https://doi.org/10.1109/LRA.2022.3180108}{DOI}.\par
\noindent [5] M. Shridhar, L. Manuelli, and D. Fox. ``CLIPort: What and Where Pathways for Robotic Manipulation.'' \emph{CoRL}, PMLR 164, 2022, pp. 894--906. \href{https://proceedings.mlr.press/v164/shridhar22a.html}{PMLR}.\par
\noindent [6] M. Shridhar, L. Manuelli, and D. Fox. ``Perceiver-Actor: A Multi-Task Transformer for Robotic Manipulation.'' \emph{CoRL}, PMLR 205, 2023, pp. 785--799. \href{https://proceedings.mlr.press/v205/shridhar23a.html}{PMLR}.\par
\noindent [7] M. Ahn et al. ``Do As I Can, Not As I Say: Grounding Language in Robotic Affordances.'' \emph{CoRL}, 2022. \href{https://arxiv.org/abs/2204.01691}{arXiv:2204.01691}.\par
\noindent [8] A. Brohan et al. ``RT-1: Robotics Transformer for Real-World Control at Scale.'' \emph{Robotics: Science and Systems}, 2023. \href{https://arxiv.org/abs/2212.06817}{arXiv:2212.06817}.\par
\noindent [9] S. Robertson and H. Zaragoza. ``The Probabilistic Relevance Framework: BM25 and Beyond.'' \emph{Foundations and Trends in Information Retrieval}, 3(4), 2009, pp. 333--389. \href{https://doi.org/10.1561/1500000019}{DOI}.\par
\noindent [10] G. Salton and C. Buckley. ``Term-Weighting Approaches in Automatic Text Retrieval.'' \emph{Information Processing \& Management}, 24(5), 1988, pp. 513--523. \href{https://doi.org/10.1016/0306-4573(88)90021-0}{DOI}.\par
\noindent [11] S. Deerwester et al. ``Indexing by Latent Semantic Analysis.'' \emph{Journal of the American Society for Information Science}, 41(6), 1990, pp. 391--407. \href{https://www.microsoft.com/en-us/research/publication/indexing-by-latent-semantic-analysis/}{Research record}.\par
\noindent [12] C. Manning, P. Raghavan, and H. Sch\"utze. \emph{Introduction to Information Retrieval}. Cambridge University Press, 2008. \href{https://nlp.stanford.edu/IR-book/}{Online edition}.\par
\noindent [13] N. Reimers and I. Gurevych. ``Sentence-BERT: Sentence Embeddings Using Siamese BERT-Networks.'' \emph{EMNLP-IJCNLP}, 2019, pp. 3982--3992. \href{https://aclanthology.org/D19-1410/}{ACL Anthology}.\par
\noindent [14] F. Pedregosa et al. ``Scikit-learn: Machine Learning in Python.'' \emph{JMLR}, 12, 2011, pp. 2825--2830. \href{https://jmlr.org/papers/v12/pedregosa11a.html}{JMLR}.\par
\endgroup
\end{document}